\PassOptionsToPackage{prologue,dvipsnames}{xcolor}
\documentclass[10pt,twocolumn,letterpaper]{article}

\usepackage[pagenumbers]{cvpr} 

\definecolor{cvprblue}{rgb}{0.21,0.49,0.74}
\usepackage[pagebackref,breaklinks,colorlinks,allcolors=cvprblue]{hyperref}

\usepackage{bm}
\usepackage{multirow}
\usepackage{colortbl}

\usepackage[accsupp]{axessibility}

\newcommand{\cmark}{\ding{51}}

\newcommand{\pub}[1]{{\color{gray}{\footnotesize{[{#1}]}}}}
\usepackage{graphicx}
\usepackage{pifont}
\usepackage{makecell}
\definecolor{aliceblue}{RGB}{240, 248, 255}
\newcommand{\dplus}[1]{\fontsize{8pt}{0.1em}\selectfont \textbf{\textcolor{blue}{(#1)}}}
\def\paperID{3146} 
\def\confName{CVPR}
\def\confYear{2025}

\title{Uncertainty-Instructed Structure Injection for Generalizable \\HD Map Construction}

\author{Xiaolu Liu$^{1}\thanks{Equal Contribution}$, \ Ruizi Yang$^1\footnotemark[1]$, \ Song Wang$^1$, \ Wentong Li$^2$, \ Junbo Chen$^3\footnotemark[2]$, \ Jianke Zhu$^1\thanks{Corresponding authors}$\\
	$^1$Zhejiang University \ \ \
	$^2$NUAA \ \ \
    $^3$Udeer.ai \ \ \
    \\ 
    {\tt\small \{xiaoluliu, ruiziyang, songw,  jkzhu\}@zju.edu.cn, wentong\_li@nuaa.edu.cn, junbo@udeer.ai}
}

\begin{document}
\maketitle
\begin{abstract}
Reliable high-definition (HD) map construction is crucial for the driving safety of autonomous vehicles. Although recent studies demonstrate improved performance, their generalization capability across unfamiliar driving scenes remains unexplored. To tackle this issue, we propose \textbf{\textit{UIGenMap}}, an uncertainty-instructed structure injection approach for generalizable HD map vectorization, which concerns the uncertainty resampling in statistical distribution and employs explicit instance features to reduce excessive reliance on training data. Specifically, we introduce the perspective-view (PV) detection branch to obtain explicit structural features, in which the uncertainty-aware decoder is designed to dynamically sample probability distributions considering the difference in scenes. With probabilistic embedding and selection, UI2DPrompt is proposed to construct PV-learnable prompts. These PV prompts are integrated into the map decoder by designed hybrid injection to compensate for neglected instance structures. To ensure real-time inference, a lightweight Mimic Query Distillation is designed to learn from PV prompts, which can serve as an efficient alternative to the flow of PV branches. Extensive experiments on challenging geographically disjoint (geo-based) data splits demonstrate that our UIGenMap achieves superior performance, with +5.7 mAP improvement on the nuScenes dataset. Source code will be available at \href{https://github.com/xiaolul2/UIGenMap}{https://github.com/xiaolul2/UIGenMap}.
\end{abstract}
    
\section{Introduction}
Reliable HD map construction is crucial for driving safety of autonomous vehicles, which provides essential geometric information for ego-localization, perceptions, path planning, and predictions~\cite{he2023egovm, li2024ego, jiang2023vad, wang2024not}. Recently, online high-definition (HD) map vectorization has attracted considerable attention for its cost-effectiveness and ability to provide real-time updates~\cite{liao2022maptr,zhou2024himap,liu2024leveraging}.
Given images captured by onboard cameras, current map construction relies on transformer scheme~\cite{zhu2020deformable} to convert perspective-view (PV) image features into bird's-eye-view (BEV) space. Then, a decoder follows to predict the map elements. 

\begin{figure}
  \centering
  \includegraphics[width=0.48\textwidth]{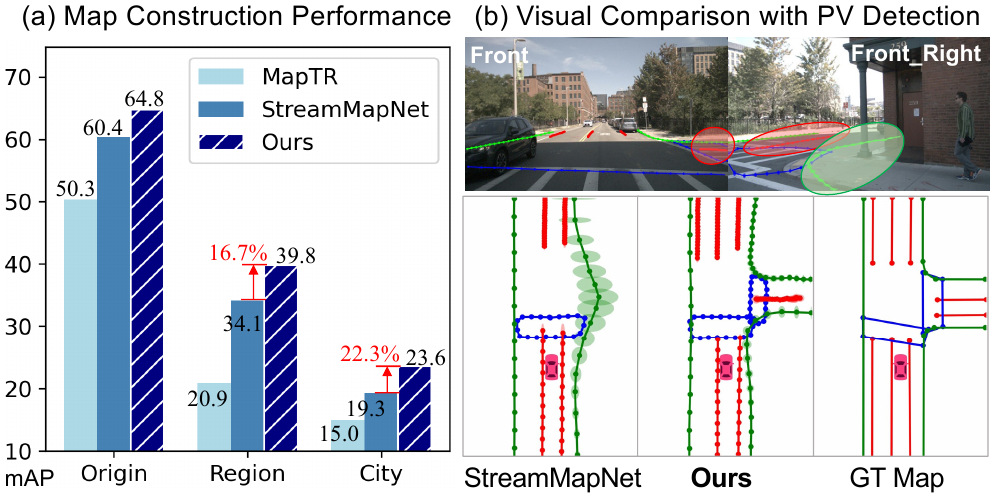}
  \vspace{-7mm}
  \caption{(a) Comparisons on original and different geo-based data partitions among previous methods and ours. (b) Visual comparisons with circles on BEV map to represent the learned uncertainties, in which PV instances serve as compensation and provide a detailed understanding of map perception with generalization.}
  \label{fig1}
  \vspace{-4mm}
\end{figure}

Despite the achievements on widely adopted benchmarks~\cite{liao2024maptrv2, liu2024mgmap, chen2025maptracker}, 
little attention has been paid to investigating the generalization capabilities, which refer to the model's performance on unfamiliar driving scenarios that are different from the training sets.
As discussed in~\cite{yuan2024streammapnet,lilja2024localization}, in the widely used public datasets~\cite{nuScenes, Argoverse2}, there are a large number of geographical overlaps between the training and validation scenes. The models are then more likely to memorize similar driving scenes rather than fully capture their underlying road structures. Variations in driving scenes, such as the contrast between bustling city centers and suburbs with simple road structures, highlight significant shifts in feature distributions and reveal the model's limitations in robustness. Besides, learning-based BEV transformation will inevitably lead to geometric errors and loss of texture details. As shown in Figure~\ref{fig1}(a), previous methods~\cite{liao2022maptr, yuan2024streammapnet} exhibit notable performance degradation on region-based and city-based data splits under changing driving scenes. 

To tackle the above issue, we propose the innovative employment of model uncertainty targeted for more generalizable HD map constructions. On the one hand, the uncertainty estimation process involves learning both the statistical mean and variance value, enabling dynamic resampling based on probabilistic distributions. Given the discrepancy in feature distribution across different driving scenes, such a statistical resampling strategy enhances the model's dynamic adaptability in unfamiliar environments. However, the estimated model uncertainty quantifies the confidence in its predictions, which can serve as a reliable indicator for further feature selection and enhancement.
As illustrated in Figure~\ref{fig1}(b), direct 2D perception can capture more intuitive semantic and angular structural information, which might be missed or incorrectly learned during the implicit PV-to-BEV conversions. Thus, combined with uncertainty-based feature selection, more reliable explicit information can be emphasized as compensation to boost the robustness of BEV-level map construction. 

Based on the above discussion, in this work, we propose \textit{\textbf{UIGenMap}}, an uncertainty-instructed PV structure injection strategy for more generalizable HD map vectorization. 
Specifically, 
a PV detection branch is introduced to provide explicit structural instances. 
In both PV and BEV branches, we move beyond the vanilla transformer structure by designing the statistically sensitive uncertainty-aware decoder (UA-Decoder), which constructs the probabilistic outputs by uncertainty-aware attention and finer point-level uncertainty head.
Leveraging the PV uncertainty output, we introduce the UI2DPrompt to construct learnable PV prompts. These PV prompts are integrated into the BEV decoder by the designed hybrid injection mechanism, which acts as structural compensation for the BEV map prediction. Consequently, reliable structural representation from PV space can be injected to enhance the model’s capability, thereby improving its perceptual ability in complex environments. Considering the demands for real-time inference, we also design a lightweight mimic query distillation module to learn from constructed PV prompts. Then during inference, the mimicked query can replace the PV detection flow to reduce extra computational burdens.

To prioritize the evaluation of generalization capability, we assess our method using geographically disjoint (geo-based) data partitions on the nuScenes~\cite{nuScenes} and Argoverse2~\cite{Argoverse2} datasets. Extensive experiments demonstrate that our UIGenMap achieves state-of-the-art performance with higher adaptive ability, especially with +5.7 mAP improvement on the region-based partition and +4.3 mAP on the more challenging city-based split on nuScenes.  
The main contributions can be summarized as follows:
\begin{itemize}
    \item  We propose UIGenMap, an uncertainty-instructed approach to more generalizable HD map vectorization with explicit structure injection.
    \item The UA-Decoder and UI2DPrompt are proposed to construct structural PV prompts, which are dynamically integrated with BEV perception by hybrid injection.
    \item Lightweight MQ-Distillation design supports real-time inference without the additional computational overhead of the PV detection branches.
    \item Promising results on more challenging geo-based data partitions show that UIGenMap outperforms previous approaches and achieves powerful generalization capability.
\end{itemize}
\section{Related Work}
\noindent \textbf{Vectorized HD Map Construction. }  
Conventional HD map construction approaches~\cite{shan2018lego,shan2020lio,zhu2024mesh} rely on SLAM-based annotations, which require tedious offline manual labeling at high costs. Recent studies ~\cite{liu2023vectormapnet, ding2023pivotnet, qiao2023end, li2024dtclmapper, li2024genmapping,yang2024mgmapnet,hu2024admap} have explored online approaches with the input of onboard sensor data. In~\cite{li2022hdmapnet,wang2023lidar2map,li2022bevformer, philion2020lift, liang2022bevfusion}, HDMap construction is treated as a segmentation task, relying on post-processing to achieve map vectorization. Later, Liao \textit{et al.}~\cite{liao2022maptr, liao2024maptrv2} represent map elements as ordered points and employ transformers~\cite{vaswani2017attention,zhu2020deformable} after the BEV encoder for auto-regressive detection. 
Other modeling methods, such as Bezier curves and pivot points, are explored in
BeMapNet~\cite{qiao2023end} and PivotNet~\cite{ding2023pivotnet} for geometric representation. 
Several works~\cite{yuan2024streammapnet,wang2024stream,li2024dtclmapper,chen2025maptracker, song2024memfusionmap} are proposed for different temporal fusion approaches. GeMap~\cite{zhang2023gemap} and MapQR~\cite{liu2024leveraging} improve mapping performance from the perspectives of geometric relationships and decoder design, respectively. In ~\cite{zhang2024enhancing,liu2024mgmap,zhang2024online,choi2024mask2map}, rasterized maps are utilized for structure compensation. Despite improved performance, these approaches seldom consider the generalization capability of models in unfamiliar driving scenarios.  In this work, we incorporate uncertainty-instructed PV detection retracing, aiming for better performance across diverse scenarios.

\noindent \textbf{PV Detection for 3D Perception.}
Instances from PV detection can be effective compensations for 3D perceptions ~\cite{yang2023bevformerv2,zhang2023simple,lu2023towards,jiang2024far3d,ji2024enhancing,wuboosting}. For 3D object detection, BEVFormerv2~\cite{yang2023bevformerv2} utilizes a two-stage BEV detector that integrates the query from the PV head into the BEV space. In SimMoD~\cite{zhang2023simple}, 2D detection is utilized to construct proposals for 3D fine-grained perception. Far3D~\cite{jiang2024far3d} combines 2D detection boxes with estimated depth information to improve long-range perception performance. 
For model generalization, Lu \textit{et al.}~\cite{lu2023towards} propose a generalizable framework based on perspective debiasing, in which PV consistency learning is employed among different domains. Applying for 3D lanes, Li \textit{et al.}~\cite{li2024enhancing} use 2D priors for lane detection and topology reasoning. Inspired by the above methods, we employ PV structural features for more generalizable HD map vectorization, in which distillations from 2D instances are utilized to ensure inference efficiency.

\begin{figure*}[th!]
\begin{center}
\centering
\includegraphics[width=0.99\textwidth]{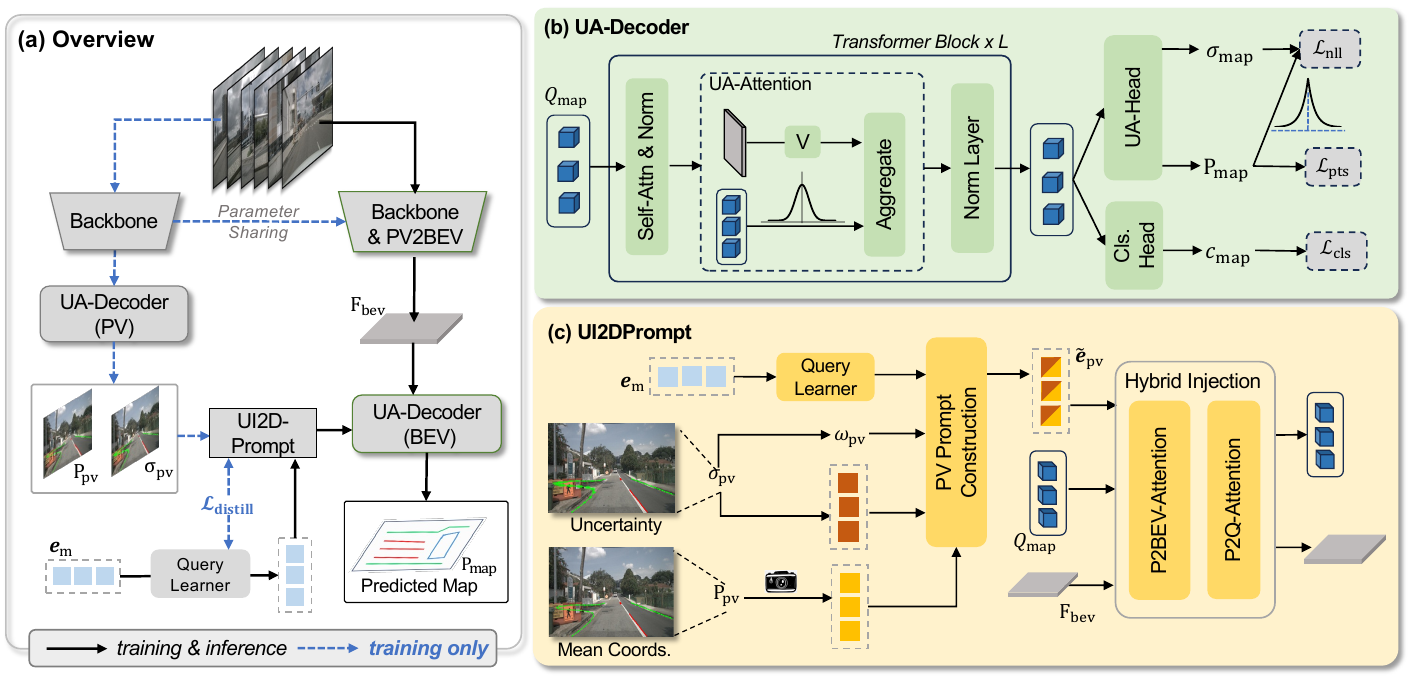}
\vspace{-3mm}
\caption{\textbf{(a) Overview of Our UIGenMap}. For training, the PV branch is introduced with the uncertainty-instructed structural injection, in which the MQ-Distillation is designed to mimic PV structural features. \textbf{(b) Uncertainty-Aware Decoder Architecture}. For each
layer, UA-Decoder comprises probabilistic UA-Attention and UA-Head for reliable output. \textbf{(c) UI2DPrompt Design.}
We construct PV prompts from PV-detected elements and their corresponding uncertainties, which are integrated with the main branch by hybrid injection.}
\label{fig_framework}
\vspace{-6mm}
\end{center}
\end{figure*} 

\noindent \textbf{Uncertainty in Model.}
 Generally, uncertainty refers to the model's confidence level for its predictions, which can be a statistical representation to indicate the model's understanding of scenes~\cite{chang2020data,kendall2017uncertainties, wang2024reliocc}. Uncertainty has gained intensive attention due to the potential for reliability evaluation and performance enhancement~\cite{lakshminarayanan2017simple, cao2024pasco}. Specifically, the generalization problem with uncertainty modeling is investigated in~\cite{qiao2021uncertainty,li2022uncertainty}, which expects uncertainty estimation for out-of-distribution generalization. For the transformer structure~\cite{pei2022transformer, guo2022uncertainty}, uncertainty has been explored to guide the training and inference processes for more robust model performance. 
In the field of autonomous driving, significant uncertainty has been introduced for 3D occupancy~\cite{cao2024pasco, wang2024reliocc}, optical flow~\cite{luo2023gaflow}, and object tracking~\cite{zhou2024ua}.
Recently, Gu \textit{et al.}~\cite{gu2024producing} present uncertainty estimation for the vectorized map, which is targeted for trajectory prediction. 
In our design, we utilize uncertainty for more generalized HD map construction, which considers the variability in feature distributions and incorporates explicit instance features. 

\section{UIGenMap}

Given surrounding-view images captured from onboard cameras, we aim to construct more robust vectorized map elements in the BEV space. Each map element comprises a class label $c_{\text{map}}$ and an ordered sequence of points $\text{P}_{\text{map}}=\{(p_x^i,p_y^i)\}_{i=1}^N$ that represents the structure of the instance. $N$ denotes the total number of points for each map element. 

The architecture of UIGenMap is presented in Figure~\ref{fig_framework}. Based on the main BEV detection, we obtain PV instance from an introduced pre-trained PV branch, which is injected as structural priors for BEV perception.
Firstly, we extract image features
for the PV and BEV branches separately. As in~\cite{li2022bevformer}, BEV features $\text{F}_{\text{bev}}$ are constructed from the interaction between PV features and learnable BEV queries. 
Then, we design the uncertainty-aware (UA-) decoder from inner-layer UA-Attention and point-level heads to achieve reliable probabilistic outputs. 
Based on the PV coordinates and uncertainty output, we propose the UI2DPrompt module to construct reliable PV prompts. These prompts are incorporated into BEV features and map instance queries via hybrid injection, where uncertainties are utilized as the basis for feature selection and enhancement. Finally, the UA-Decoder employs enhanced BEV features and instance queries for reliable map predictions. To ensure real-time inference, the lightweight MQ-Distillation module performs query distillation from learned PV prompts to imitate explicit structural features, aiming for efficient computation.

\subsection{Uncertainty-Aware Decoder}
\label{ua-decoder}

We design UA-Decoder for both PV and BEV space.  
Systematic uncertainty-aware design is employed to improve the statistical understanding of features and outputs to achieve more dynamic self-adjustment for different driving scenarios. To ensure comprehensive perception, each decoder layer comprises UA-Attention at the instance feature level and UA-Head for point-level uncertainty output along with positions, respectively.

\begin{figure}
  \centering
  \includegraphics[width=0.48\textwidth]{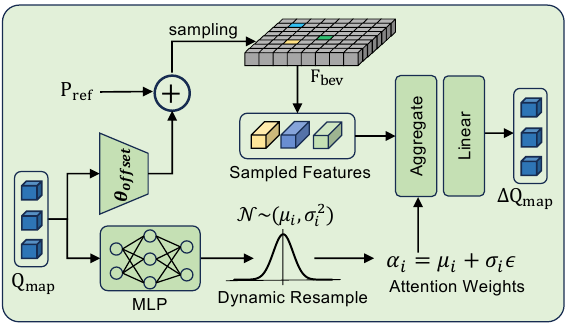}
    \vspace{-7mm}
  \caption{UA-Attention Design. The learned probabilistic weights $\alpha_i$ are dynamically resampled from Gaussian distribution with stochastic disturbance, which is multiplied with sampled feature values to update map queries.}
   \vspace{-5mm}
   \label{uattention}
\end{figure}

\noindent{\textbf{UA-Attention.}} In conventional deformable attention~\cite{zhu2020deformable}, the weight $\alpha_i$ for each sampled feature values from BEV feature $F_{\text{BEV}}$ is determined by linear layer ($\texttt{Linear}(\cdot)$) on $Q_{\text{map}}^i$, which can be expressed as $\alpha_i = 
\texttt{Linear}(Q_{\text{map}}^i)$, where $Q_{\text{map}}^i$ is the $i$-th instance query of totally $M$ map elements. Then, the updated part of query $\Delta Q_{\text{map}}^i$ is aggregated by the dot-product between the learned weight $\alpha_i$ and sampled feature values from $\text{F}_{\text{bev}}$. Considering complex driving scenarios, 
such deterministic weights are inadaptable for challenging situations.
More dynamic weights are required to avoid potential ambiguities. Inspired by~\cite{pei2022transformer, guo2022uncertainty}, we explicitly model the attention scores under a Gaussian distribution $\alpha_i\sim\mathcal{N}(\mu_i,\sigma_i^2)$, as shown in Figure~\ref{uattention}.  We adopt the re-parameterization strategy to perform the probabilistic attention, which is constructed by the corresponding mean $\mu_i$ and variance $\sigma_i$.  Such probabilistic parameters are predicted through multilayer perception ($\texttt{MLP}(\cdot)$). Thus, the construction of dynamic resampled attention weights can be formulated as below:
\begin{align}
\{\mu_i,\sigma_i^2\} = \texttt{MLP}(Q_{\text{map}}^i),\\
\alpha_i=\mu_i+\sigma_i\epsilon, \epsilon\sim\mathcal{N}(0,1),
\end{align}
\noindent where $\epsilon$ is the Gaussian noise conforming to normal distribution.
During inference, attention
 weight $\alpha_i$ can be dynamically resampled from learned statistical distribution to suit for driving scenarios with large variations. Then query features can be obtained by dot production with the sampled feature values in order to enable reliable query updates.

\noindent{\textbf{UA-Head with Uncertainty Loss.}} Based on the robust design of UA-Attention, quantized uncertainty for each lane point output is also essential, which is point-specific at a fine-grained level to explicitly reflect the reliability and uncertainty information within the internal transformer layers. 
Inspired by~\cite{gu2024producing},  we replace the vanilla deterministic output with the UA-Head that predicts coordinates $\text{P}_{\text{map}}=\{(\hat{p}^i_x,\hat{p}^i_y)\}^N_{i=1}$ 
 along with their corresponding uncertainties $\sigma_{\text{map}}=\{(\sigma^i_x, \sigma^i_y )\}^N_{i=1}$ for each map element, as shown in Figure~\ref{fig_framework} (b). These parameters represent the location and scale of the joint Laplace distribution $Laplace(\hat{p}^i_j, \frac{\sigma^i_j}{\sqrt{2}})$ at the $j$-th dimension ($x$ or $y$) of the $i$-th point, which refers to a probabilistic representation for stable output.

For uncertainty-aware training, we combine probabilistic Negative Log-Likelihood (NLL) loss $\mathcal{L}_{\text{nll}}$ with point regression loss $\mathcal{L}_{\text{pts}}$ in~\cite{yuan2024streammapnet} to balance uncertainties and coordinate outputs, in which $\mathcal{L}_{\text{nll}}$ can be formulated as follows:
\begin{equation}
\mathcal{L}_{\text{nll}} = \lambda_{\text{nll}}\frac{1}{N} \sum_{i=1}^{N}\sum_{j=1}^{2} \left( \log(2\sigma^i_j) + \frac{|p^i_j - \hat{p}^i_j|}{\sigma^i_j} \right),
\end{equation}
 with $\lambda_{\text{nll}}$ as the loss weight setting.

As illustrated in Figure~\ref{fig_framework} (a), UA-Decoder is used for the construction of the PV branch and the BEV map. To provide more robust structural features in PV branch, PV coordinates $\text{P}_{\text{pv}}=\{x^i_{\text{pv}},y^i_{\text{pv}}\}_{i=1}^N$ and the uncertainty parameters $\sigma_{\text{pv}}=\{\sigma^i_{\text{pv,x}},\sigma^i_{\text{pv,y}}\}_{i=1}^N$ are constructed by the UA-Decoder structure, which is later utilized in our UI2DPrompt design.
Thus, the systematic combination of uncertainty-aware design provides dynamic sampling during the learning process, along with established uncertainty, to construct a reliable basis for confidence scoring. 
\subsection{UI2DPrompt} 
\label{ui2dprompt}
PV branches contain abundant structural and geometric details. To mitigate information loss and misinterpretation during query-based PV-to-BEV transformation, reliable structural PV instances can be injected as compensation for BEV prediction. This section focuses on constructing PV prompts based on uncertainty-aware PV predictions, which are dynamically integrated with the map decoder for BEV map element detection.

\noindent\textbf{Prompt Construction.}  Given the learned PV instance coordinates and the corresponding uncertainty parameters $\{\text{P}_{\text{pv}}, \sigma_{\text{pv}}\}$, PV prompts are constructed to provide dynamic 2D priors. Firstly, we employ the estimated classification score $c_\text{map}$ for PV candidates selection. When $c_\text{map}>c_{\text{thr}}$, PV instances are chosen as candidates for reliable localized information. Considering the sparse depth information and the characteristics of static map elements on the ground, we utilize Inverse Perspective Mapping (IPM) to transform $\text{P}_{\text{pv}}$ from the PV image coordinate to the BEV coordinate system $\text{P}_{\text{trans}}=\{p'_x, p'_y\}$, which makes a flat grounding assumption and utilizes the affine transformation between the PV and BEV coordinates. Given the corresponding intrinsic parameters $\text{K}_{\text{cam}}$ and extrinsic transformation matrix $\text{T}_{\text{ego2cam}}$, the projection $\text{P}_{\text{pv}}$ can be defined as:
\begin{equation}
\text{P}_{\text{trans}} = \texttt{Proj}_{\texttt{2D}}\left( \text{T}_{\text{ego2cam}}^{-1} \cdot \begin{bmatrix} \text{K}_{\text{cam}}^{-1} \cdot \text{P}_{\text{pv}} \\ 1 \end{bmatrix} \right),
\end{equation}
where $\texttt{Proj}_{\texttt{2D}}(\cdot)$ denotes the conversion from homogeneous coordinates into 2D plane coordinates. 

To ensure reliable enhanced features, the corresponding uncertainty parameters $\sigma_{\text{pv}}$ cooperate as weights and feature embeddings. We encode the transformed coordinates $\text{P}_{\text{trans}}$ and uncertainty parameters $\sigma_{\text{pv}}$ by $\phi_p(\cdot)$ and $\phi_{\sigma}(\cdot)$, respectively. Point-specific embeddings $\{\bm{e}^i_{pv}\}_{i=1}^{N}$ are constructed by concatenation:
\begin{equation}
    \bm{e}^i_{\text{pv}} = \texttt{Concat}([\phi_p(\text{P}_{\text{trans}}), \phi_{\sigma}(\sigma_{\text{pv}})]).
\end{equation}
Thus, the structural PV output coordinates and uncertainty information are dynamically encoded. 

Based on the above, PV uncertainties can serve as the quantized weights for dynamic feature strengthening and adjustment. Reliable confidence indicator is able to be formulated as the weight $\{\omega^i_{\text{pv}}\}_{i=1}^N$ concerning each point-specific prompt. Then, the enhanced PV prompt $\tilde{\bm{e}}^{i}_{\text{pv}}$ can be constructed by:
\begin{equation}
    \omega^i_{\text{pv}} = \mathrm{\exp}\Bigg (\frac{(\Vert\sigma^i_{\text{pv}} \Vert_2)^{-1}}{\sum_{i=1}^N (\Vert \sigma^i_{\text{pv}} \Vert_2)^{-1}}\Bigg), \quad \tilde{\bm{e}}^{i}_{\text{pv}} = \omega^i_{\text{pv}} \bm{e}^i_{\text{pv}}+\bm{e}^i_{\text{m}}, \label{w_e}
\end{equation}
where $\exp(\cdot)$ represents the exponential amplification of the normalized inverse uncertainty value. $\{\bm{e}^i_{\text{m}}\}_{i=1}^N$ is the predefined mimic query, which is utilized for MQ-Distillation. This approach reduces the introduction of PV errors, ensuring that injecting PV instances amplifies reliable PV information while minimizing uncertainty-induced errors.

\noindent\textbf{Hybrid Injection.} 
Constructed from PV instances with uncertainty output, PV prompts $\{\tilde{\bm{e}}^{i}_{\text{pv}} \}_{i=1}^N$ are incorporated as compensation for the construction of the BEV map. These prompts are injected into the BEV feature $\text{F}_{\text{bev}}$ and the map queries $Q_{\text{map}}^i$ for the map decoder by the designed hybrid injection mechanism, which includes both prompt-to-BEV (P2BEV) and prompt-to-query (P2Q) attention.

\begin{figure}[t!]
  \centering
  \includegraphics[width=0.48\textwidth]{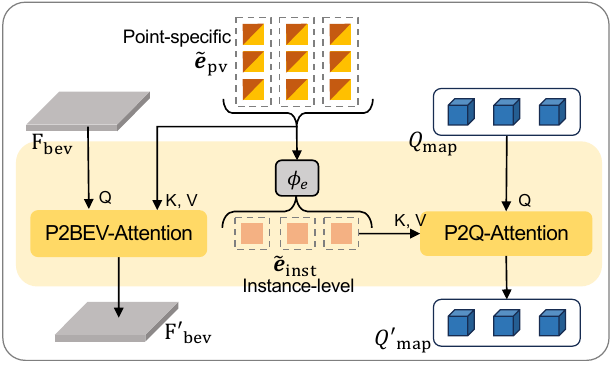}
\vspace{-7mm}
  \caption{Design of Hybrid Injection. P2BEV-Attention and P2Q-Attention are designed for PV prompts to integrate with BEV features and map instance queries as structural compensation.
  }
  \vspace{-4mm}
   \label{hybrid_injection}
\end{figure}

As shown in Figure~\ref{hybrid_injection}, point-level PV prompts $\{\tilde{\bm{e}}^{i}_{\text{pv}}\}_{i=1}^N$ are integrated with $\text{F}_{\text{bev}}$ through cross-attention of P2BEV. This step enriches BEV features by integrating detailed, point-specific information from PV prompts, enhancing instance-specific detail in the BEV representation:
\begin{equation}
    \text{F}'_{\text{bev}}=\texttt{Attn}(W_q \text{F}_{\text{bev}},\; W_k \tilde{\bm{e}}^{i}_{\text{pv}},\; W_v \tilde{\bm{e}}^{i}_{\text{pv}}),
\end{equation}
where $W_{\{q,k,v\}}$ are the corresponding projection matrix. Then, we encode point-level prompts into instance-level representations through a linear block $\phi_e(\cdot)$, which aggregates point-specific information to create a higher-level representation for subsequent processing. These instance-level prompts $\tilde{\bm{e}}_{\text{inst}}$ are used to integrate with $Q_{\text{map}}$, which inject reliable priors for the initialized queries and enhance the overall map construction:
\begin{equation}
    \tilde{\bm{e}}_{\text{inst}} = \phi_e\left(\{\tilde{\bm{e}}^{i}_{\text{pv}}\}_{i=1}^N\right),
\end{equation}
\begin{equation}
    Q'_{\text{map}} = \texttt{Attn}\left(W_q  Q_{\text{map}},\; W_k  \tilde{\bm{e}}_{\text{inst}}, \; W_v \tilde{\bm{e}}_{\text{inst}}\right).
\end{equation}
By injecting PV priors into both BEV and decoder queries, this hybrid mechanism improves the reliability of BEV perception and reduces transition errors. Explicit structural features enhance the generalization capability of the learned model and perform robustly in various driving environments, ensuring better accuracy and consistency.

 \begin{table*}[ht!]
\begin{center}
\centering
\renewcommand\arraystretch{1.1}
\setlength{\tabcolsep}{7pt}
\resizebox{0.99\textwidth}{!}{
\begin{tabular}[t]{r|cc|cccc|cccc|c}
\toprule[1pt]

  & &  &\multicolumn{4}{c|}{\textbf{Region-Based}} & \multicolumn{4}{c|}{\textbf{City-Based}} &  \\
\multicolumn{1}{c|}{\multirow{-2}{*}{\textbf{Method}}}& \multirow{-2}{*}{\textbf{Backbone}} & \multirow{-2}{*}{\textbf{Epochs}}  & \textbf{AP}$_{\mathbf{ped}}$ & \textbf{AP}$_{\mathbf{div}}$ & \textbf{AP}$_{\mathbf{bou}}$ & \textbf{mAP} & \textbf{AP}$_{\mathbf{ped}}$ & \textbf{AP}$_{\mathbf{div}}$ & \textbf{AP}$_{\mathbf{bou}}$ & \textbf{mAP} & \multirow{-2}{*}{\textbf{FPS}}\\

\midrule
VectorMapNet*~\pub{ICML'23} & R50& 120 & 15.8 &17.0&21.2& 18.0 &-&-&- & -  &  -\\
MapTR*~\pub{ICLR'23} & R50& 24 & 6.4 &20.7&35.5& 20.9 &11.1&12.9&21.1 & 15.0  & \textbf{15.8} \\
MGMap~\pub{CVPR'24} &R50 & 24 & 9.4 & 27.0  & 38.1 &24.8 & 10.9  &17.8  & 24.7 & 17.8   & 11.5 \\
MapQR~\pub{ECCV'24}  & R50& 24 &8.4 & 25.7 & 39.9 &  24.7 & 10.2  &15.8  & 25.5 & 17.2 & 12.1 \\
GeMap~\pub{ECCV'24} & R50& 24 & 11.8  & 26.6 & 43.6 &  27.3 & 12.9&16.3& 26.7 & 18.6   &  11.6 \\
MapTRv2*~\pub{IJCV'24} &R50 & 24 &  14.7 & 28.7 & 43.3 &  28.9 &18.9& 18.6 & \textbf{27.9} &21.8    & 12.9 \\
StreamMapNet*~\pub{WACV'24} & R50 & 24 &  \underline{32.2} & 29.3 & 40.8 & 34.1  &18.7& 16.4& 22.7 & 19.3    & 13.3 \\
\rowcolor{blue!10}
\textbf{\textit{UIGenMap-d (Ours)}} & R50& 24 & \textbf{40.3} & \underline{30.8} &\underline{46.8} &  \underline{39.3} \dplus{+5.2} &  \underline{23.1}&\underline{19.6} &25.4 & \underline{22.7} \dplus{+3.4}&  12.2\\
\rowcolor{blue!10}
\textbf{\textit{UIGenMap (Ours)}} & R50& 24 & \textbf{40.3} & \textbf{31.8} & \textbf{47.2} & \textbf{39.8} \dplus{+5.7} & \textbf{24.1}  & \textbf{20.7} &  \underline{26.1}& \textbf{23.6} \dplus{+4.3}&  8.2 \\

\midrule
StreamMapNet~\pub{WACV'24} &SwinT  & 24 & 32.5 &  27.4 & 41.7&  33.9 & 21.8 & 19.7 & 25.8 & 22.4 & 9.3\\

\rowcolor{blue!10}
\textbf{\textit{UIGenMap-d (Ours)}} & SwinT& 24 & \textbf{42.6} & \textbf{31.9} & \textbf{47.2}& \textbf{40.6} \dplus{+6.7}  &\textbf{24.7} & \textbf{21.2} & \textbf{28.8} & \textbf{24.9} \dplus{+2.5}&8.5\\

\bottomrule[1pt]

\end{tabular}}
\end{center}
\vspace{-6mm}
\caption{Quantitative comparisons on \textit{nuScenes} validation dataset, utilizing region-based and city-based dataset partitions. 
Methods with ``*'' denote that data results are obtained from corresponding papers~\cite{yuan2024streammapnet, lilja2024localization}, and others are reproduced by official public codes. ``R50'' and ``SwinT'' denotes the image backbone of ResNet50~\cite{he2016deep} and Swin-Tiny~\cite{liu2021swin}. ``\textit{UIGenMap-d}'' represents the utilization of mimic query from MQ-Distillation. FPS is measured on a single RTX A40 GPU.} 
\vspace{-4mm}
\label{exp_nus}
\end{table*}

\subsection{Mimic Query Distillation}
To address the extra computation introduced by the PV branch, we propose the lightweight MQ-Distillation to distill from PV prompt construction, where the mimic query can serve as an alternative to achieve real-time inference.

In MQ-Distillation, we define a set of mimic queries $\{\bm{e}^i_{m}\}_{i=1}^{N}$, along with an MLP-based query learner $h(\cdot)$, which is designed to capture and distill the structural features of the PV prompt construction process $\tilde{\bm{e}}^{i}_{\text{pv}}$. 
As shown in Equation~\ref{w_e}, during training, the predefined mimic queries $\bm{e}^i_{m}$ and the constructed PV prompts $\tilde{\bm{e}}^{i}_{\text{pv}}$ are jointly combined by addition, which are sent to the hybrid injection for further operations. 

To obtain sufficient structural features and achieve progressive alignment with PV prompts, we design the loss function that guides mimic queries for distillation.
We use Mean Squared Error (MSE) as the distillation loss to measure the difference between the constructed PV prompts and the mimicked queries,
which can be formulated as follows: 
\vspace{-1mm}
\begin{equation} 
\mathcal{L}_{\text{distill}} = \lambda_{\text{distill}} \frac{1}{N} \sum_{i=1}^{N} \left(\tilde{\bm{e}}^{i}_{\text{pv}} - h(\bm{e}^i_m)\right)^2, \label{eq} 
\end{equation} 
where $\lambda_{\text{distill}}$ is a weight coefficient that balances the impact of the distillation loss on the overall training process. Such loss function guides the mimic queries to accurately replicate the PV prompts generated by the PV branch, making them reliable substitutes for inference.

To this end, only the mimic queries can be utilized during inference, which eliminates the requirement for PV prompt computation. This approach maintains real-time inference capability by efficiently mimicking PV structural features with minimal computational cost, enabling streamlined yet robust map construction.

\subsection{Training and Inference}

During the training process, combined with $\mathcal{L}_{\text{nll}}$ for uncertainty output and $\mathcal{L}_{\text{distill}}$ for mimic query distillation, the overall loss setting $\mathcal{L}_{\text{map}}$ can be represented as:
\begin{equation}
\mathcal{L}_{\text{map}} = \lambda_1\mathcal{L}_{\text{pts}} + 
\lambda_2\mathcal{L}_{\text{cls}}+\mathcal{L}_{\text{nll}} + \mathcal{L}_{\text{distill}},
\end{equation}
where $\mathcal{L}_{\text{pts}}$ denotes the point-to-point Manhattan distance between the regressed points and the ground-truth. $L_{\text{cls}}$ represents the focal loss for map classification. $\lambda_1$ and $\lambda_2$ are the corresponding loss weights for point regression and map classification, respectively.

For inference, the well-learned mimic query can only be employed as a substitute for PV prompts, avoiding the computational burden for the PV detection branch. Thus, our approach ensures real-time inference without introducing unnecessary computational overhead, while preserving effective performance. Additionally, the model’s uncertainty learning enables dynamic sampling during inference, enhancing the adaptability to diverse driving environments.

 \begin{table*}[th!]
\begin{center}
\centering
\renewcommand\arraystretch{1.1}
\setlength{\tabcolsep}{7pt}
\resizebox{0.99\textwidth}{!}{
\begin{tabular}[t]{r|cc|cccc|cccc|c}
\toprule[1pt]

  & &  &\multicolumn{4}{c|}{\textbf{Region-Based}} & \multicolumn{4}{c|}{\textbf{City-Based}} &  \\
\multicolumn{1}{c|}{\multirow{-2}{*}{\textbf{Method}}}& \multirow{-2}{*}{\textbf{Backbone}} & \multirow{-2}{*}{\textbf{Epochs}}  & \textbf{AP}$_{\mathbf{ped}}$ & \textbf{AP}$_{\mathbf{div}}$ & \textbf{AP}$_{\mathbf{bou}}$ & \textbf{mAP} & \textbf{AP}$_{\mathbf{ped}}$ & \textbf{AP}$_{\mathbf{div}}$ & \textbf{AP}$_{\mathbf{bou}}$ & \textbf{mAP} & \multirow{-2}{*}{\textbf{FPS}}\\

\midrule
VectorMapNet*~\pub{ICML'23} & R50 & 120 & 35.6 & 34.9 & 37.8 &  36.1 &  -&- &- & -   & -\\
MapTR*~\pub{ICLR'23} & R50 & 30 & 48.1 & 50.4 & 55.0 &  51.1 &  34.7 & 41.7 & 34.7 & 37.9   & \textbf{15.8} \\
GeMap~\pub{ECCV'24}& R50& 30 & 52.5  & \textbf{59.2}  & 58.2 & 56.6  & 34.5  & 41.1 & 40.4 & 38.7  &  11.6 \\
MapTRv2*~\pub{IJCV'24} & R50& 30 & 53.3  & 58.7  & 58.8 & 56.9  &37.4 & 42.2 & 41.9 & 40.5 & 12.9 \\
MapQR~\pub{ECCV'24}  & R50& 30 & 54.2  &  58.0 & 60.4 & 57.5  & 33.8  & 41.5  &41.6   & 38.9  & 12.1\\
StreamMapNet*~\pub{WACV'24} &R50  & 30 & 57.9 &  55.7 & 61.3&  58.3 & 40.2 & 43.4 & 43.3 & 42.3  & 13.3\\
\rowcolor{blue!10}
\textbf{\textit{UIGenMap-d (Ours)}} &R50 & 30 & \underline{58.3}  &  58.4 & \underline{62.4}& \underline{59.7} \dplus{+1.4}& \underline{42.9} &\underline{44.4}  &\underline{44.7} &  \underline{44.0} \dplus{+1.7}& 12.2\\  

\rowcolor{blue!10}
\textbf{\textit{UIGenMap (Ours)}} &R50 & 30 &   \textbf{59.8}&  \underline{58.7} &  \textbf{62.7}&  \textbf{60.4} \dplus{+2.1}&  \textbf{43.7} & \textbf{45.7} & \textbf{45.1} &  \textbf{44.8} \dplus{+2.5} &8.2 \\  
\bottomrule[1pt]

\end{tabular}}
\end{center}
\vspace{-6mm}
\caption{Quantitative comparisons on \textit{Argoverse2} validation datasets, utilizing region-based and city-based dataset partitions.
Methods with ``*'' denote that data results are obtained from corresponding papers~\cite{yuan2024streammapnet, lilja2024localization}, and others are reproduced by official public codes. Models are trained for 30 epochs and follow the same setting. FPS is measured on a single RTX A40 GPU.} 
\vspace{-4mm}
\label{exp_argo}
\end{table*}

\begin{table}[th!]
  \centering
  \renewcommand\arraystretch{1.1}
  \renewcommand\tabcolsep{4pt}
  \resizebox{0.48\textwidth}{!}{
\begin{tabular}{c|cc|ccc|c}
 \toprule[1pt]
\textbf{Dataset}                     & \textbf{Method }             & \multicolumn{1}{l|}{\textbf{Epochs}} & \textbf{AP}$_{\mathbf{ped}}$ & \textbf{AP}$_{\mathbf{div}}$ & \textbf{AP}$_{\mathbf{bou}}$ & \textbf{mAP} \\ \midrule
\multirow{3}{*}{nuScenes}   
&MapTRV2~\cite{liao2024maptrv2}& 24 &59.8&62.4&62.4& 61.5 \\
&StreamMapNet~\cite{yuan2024streammapnet}        & 24 &60.4&61.9&58.9& 60.4 \\
& SQD-MapNet~\cite{wang2024stream}& 24&63.0&65.5&63.3 & 63.9 \\
& \cellcolor{blue!10}\textit{\textbf{UIGenMap(Ours)}} & \cellcolor{blue!10}24 &\cellcolor{blue!10}64.2 & \cellcolor{blue!10}65.4 & \cellcolor{blue!10}64.9 & \cellcolor{blue!10}\textbf{64.8} \\ 
\midrule
\multirow{3}{*}{Argoverse2}  
&StreamMapNet~\cite{yuan2024streammapnet}        & 30&62.9&59.5& 63.0  & 61.5 \\
& SQD-MapNet~\cite{wang2024stream}& 30 &64.9&60.2&64.9 & 63.3\\
& \cellcolor{blue!10}\textit{\textbf{UIGenMap(Ours)}} & \cellcolor{blue!10}30&\cellcolor{blue!10}64.5&\cellcolor{blue!10}60.8 &\cellcolor{blue!10}66.1 &\cellcolor{blue!10}\textbf{63.8}          \\  
\bottomrule[1pt]
\end{tabular}}
 \vspace{-2mm}
\caption{Performance on nuScenes and Argoverse2 datasets with the original dataset partition at 30m range.}
  \label{exp_origin}
  \vspace{-3mm}
\end{table}

\begin{table}[th!]
  \centering
  \renewcommand\tabcolsep{3pt}
  \footnotesize
  \begin{tabular}{ ccc|cc c c}
    \toprule[1pt]
    \textbf{bkb}& \textbf{UI2DPrompt} & \textbf{UA-Decoder}& \textbf{AP}$_{\mathbf{ped}}$ & \textbf{AP}$_{\mathbf{div}}$ & \textbf{AP}$_{\mathbf{bou}}$ & \textbf{mAP} \\ 
    \midrule
     & &   & 32.2 & 29.3 & 40.8 &  34.1                 \\
    & \cmark &   & 39.9 & 28.6 & 43.8 &37.4             \\ 
     
     &  & \cmark & 36.1 &  31.3 & 44.0  &  37.1   \\
    & \cmark & \cmark  & 41.6 & 29.7  & 44.8  & 38.7  \\
    \cmark& \cmark &  & 41.4 & 29.8  & 47.4  & 39.5    \\
   \cmark &  & \cmark  & 39.2 & 30.1& 46.2 & 38.5      \\
   \rowcolor{blue!10}
     \cmark&\cmark & \cmark  & 40.3 & 31.8 & 47.2 & \textbf{39.8}  \\
    \bottomrule[1pt]
    \end{tabular}
\vspace{-2mm}
  \caption  {Ablations on the impact of component designs, `bkb' means the shared image backbone from PV branches.}
  \vspace{-5mm}
  \label{ab_comp}
\end{table}

\section{Experiments}

\subsection{Datasets and Metrics}
\noindent{\textbf{Datasets.}}
We conduct extensive experiments on widely used autonomous driving datasets, including nuScenes~\cite{nuScenes} and Argoverse2~\cite{Argoverse2}. nuScenes dataset comprises 1,000 driving scenes from Boston and Singapore. Each scene lasts approximately 20 seconds with a 2Hz sampling rate. For each sample, there are six surrounding images and 32-beam LiDAR point clouds. Argoverse2 contains 1,000 scenes from 6 cities, with each scene capturing 15 seconds of data at 20Hz from 7 cameras and 10Hz LiDAR sweeps. In our experiments, the sampling frequency for Argoverse2 is unified into 2Hz as in StreamMapNet~\cite{yuan2024streammapnet}.

\noindent{\textbf{Geo-based Data Partitions.}}
Considering the overlap of widely used training and validation partitions, we adopt geographically disjoint (geo-based) dataset partitions as benchmarks to evaluate the generalization capability of the presented model. Experiments are conducted on both \textbf{region-based} and \textbf{city-based} data partitions. The region-based partition is adopted from~\cite{yuan2024streammapnet}, where the driving locations geographically disjoin the training and validation sets. In the city-based data partition, we apply city-wise splits in~\cite{lilja2024localization}, where the training and validation data are from different cities with a larger distribution change.
On nuScenes, Boston and Onenorth are used for training, while Queenstown and Holland Village are used for validation. For Argoverse2, Miami and Pittsburgh are used in training, and the remains are employed for validation. For different dataset partitions, models are separately trained on that split's training set and evaluated on their corresponding validation set.

\noindent{\textbf{Evaluation Metrics.}}
In line with the existing works~\cite{yuan2024streammapnet,liao2024maptrv2}, we conduct experiments on three static map elements, namely \textit{lane dividers}, \textit{pedestrian crossings}, and \textit{road boundaries}. Perception range is set to $(\left[-15.0m, 15.0m\right])$ for X-axis and $(\left[-30.0m, 30.0m\right])$ for Y-axis. We use average precision as the evaluation metric, which calculates the Chamfer distance between predictions and ground truths. The mean Average Precision (mAP) is obtained by averaging AP across the distinct thresholds: $\{0.5m, 1m, 1.5m\}$.

\subsection{Implementation Details}
For fair comparisons, we employ ResNet50~\cite{he2016deep} as the backbone. As in~\cite{yuan2024streammapnet}, the size of BEV features is set to $50\times 100$. The number of map queries and query dimensions are set to $100$ and $512$. We use AdamW optimizer with a learning rate of $4e^{-4}$. Training epochs are limited to prevent potential overfitting. All models are trained on $4$ NVIDIA Tesla V100 GPUs with a batch size of $4$. The weighted factors $\lambda_1$ and $\lambda_2$ are set to $50.0$ and $5.0$, separately. $\lambda_{\text{nll}}$ and $\lambda_{\text{distill}}$ are set to $0.05$ and $10.0$. We adopt a pre-trained PV detection branch for our UIGenMap training process. 
More details are provided in the \textit{supplementary materials}.

\subsection{Main Results}

\begin{figure*}[ht!]
\begin{center}
\centering
\includegraphics[width=0.8\textwidth]{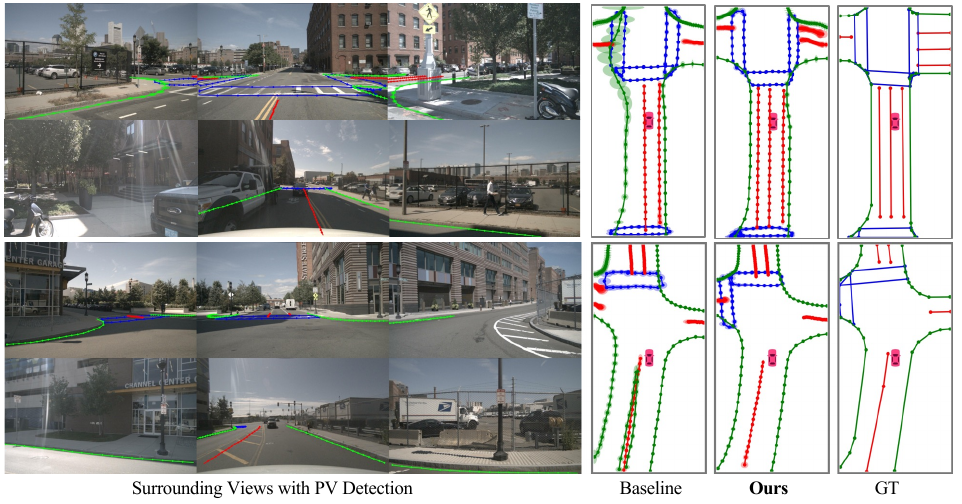}
\vspace{-3mm}
\caption{Qualitative results on the nuScenes dataset with the region-based data partition. Uncertainty outputs are represented as circles. }
\vspace{-6mm}
\label{fig_visualization}
\vspace{-2mm}
\end{center}
\end{figure*}

\noindent\textbf{Results on nuScenes.}
We compare the generalization capability with previous methods~\cite{liao2022maptr,liao2024maptrv2, liu2024leveraging,liu2024mgmap,zhang2023gemap, yuan2024streammapnet} on geospatial data partitions, in which the image size is set to 480$\times$800. As shown in Table~\ref{exp_nus}, our UIGenMap outperforms previous methods at a large margin. 
For the ResNet50 backbone, UIGenMap achieves 5.7 mAP improvement on the region-based partition, and 4.3 mAP gain is obtained on the city-based splits. 
For more efficient design, UIGenMap-d also archives 39.3 mAP with a faster inference speed. Experiments on the Swin-T~\cite{liu2021swin} backbone show significant performance with 40.6 mAP achievement.
Furthermore, we show some visual examples in Figure~\ref{fig_visualization}, indicating more intuitive achievements for UIGenMap.

\noindent\textbf{Results on Argoverse2.} More experiments are conducted on Argoverse2 dataset. The image size is 608$\times$608, and the sampling rate is 2Hz. For a fair comparison, the models are trained for 30 epochs. As shown in Table~\ref{exp_argo}, UIGenMap achieves the best results of 60.4 mAP on region-based data partitions and 43.4 mAP on city-based splits, which performs better than previous approaches. Upon further investigation, calibration errors and dataset distribution also affect model performance. More visualization can be found in our \textit{supplementary materials}. 

\noindent\textbf{Results on Original Data Partitions.} For thorough evaluation, we conducted experiments on the original data partitions with similar driving scenarios, which are widely used in previous studies. As shown in Table~\ref{exp_origin}, we compare our methods with others utilizing the same baseline. UIGenMap still outperforms the existing approaches, which outperforms StreamMapNet with +4.4 mAP and +2.3 mAP separately, validating the effectiveness of our approach.

\subsection{Ablation Studies}
In this section, ablations are conducted to verify the effectiveness of our designs. We perform experiments on nuScenes with region-based partition using ResNet50 backbone. All models are trained for 24 epochs. More ablations are provided in the \textit{supplementary materials}.

\noindent\textbf{Contributions of Main Components.} Table~\ref{ab_comp} demonstrates the impact of each component in our design. Based on the pre-trained PV backbone, we design the UA-Decoder for more reliable uncertainty output and UI2DPrompt to dynamically inject PV structural information,  which achieves +2.7 mAP and +2.4 mAP improvement, respectively. The combination of these designs gains +4.0 mAP total. The ablations are conducted on UA-Decoder at BEV space. Besides, the pre-trained PV backbone further emphasizes the effectiveness to 39.8 mAP for better scene understanding.

\noindent\textbf{Design of UA-Decoder.}
We design the UA-Attention for probabilistic attention and the UA-Head to construct the uncertainty output from the instance and finer point levels. Table~\ref{ab_uadecoder} shows the ablation without other modules. Our uncertainty-aware design supports dynamic adaptation to achieve more robust performance, which achieves +3.0 mAP improvements based on systematic combination. 

\begin{table}[th!]
  \centering
   \renewcommand\tabcolsep{3.pt}
   \footnotesize
  \begin{tabular}{ cc|cc c c}
    \toprule[1pt]
    
    \textbf{UA-Att.}& \textbf{UA-Head} & \textbf{AP}$_{\mathbf{ped}}$ & \textbf{AP}$_{\mathbf{div}}$ & \textbf{AP}$_{\mathbf{bou}}$ & \textbf{mAP} \\ 
    \midrule
    &  & 32.2 & 29.3  & 40.8 & 34.1  \\
     \cmark&     & 31.7 & 30.4 & 42.9 &  35.0    \\ 
     & \cmark   &34.8  & 30.5  & 43.4  &  36.3     \\
     \rowcolor{blue!10}
   \cmark & \cmark &  36.1  & 31.3  & 44.0 &  \textbf{37.1}  \\
    \bottomrule[1pt]
    \end{tabular}
\vspace{-2mm}
  \caption{Ablations on UA-Decoder without other modules. UA-Att and UA-Head represent the utilization of uncertainty-aware attention and output head respectively.}
  \vspace{-3mm}
  \label{ab_uadecoder}
\end{table}

\begin{table}[ht!]
\centering
\renewcommand\tabcolsep{18pt}
  \renewcommand\tabcolsep{7.pt}
   \footnotesize
  \begin{tabular}{c|c|c}
    \toprule[1pt]
    \textbf{Coords. Trans. }&\textbf{Unc. Utilization}&  \textbf{mAP}\\ 
    \midrule
    / & / &   37.6\\
     IPM& embedding only&  39.3 \\
     
    \cellcolor{blue!10}IPM & \cellcolor{blue!10} weight \& embedding & \cellcolor{blue!10}\textbf{39.8} \\
    3D Trans.& weight \& embedding & 38.7 \\
    \bottomrule[1pt]
    \end{tabular}
\vspace{-2mm}
  \caption{Ablations on the construction of PV prompts. ``3D Trans.'' means the depth-based transformation. The first row with ``/'' indicates direct PV coordinate embedding.}
  \vspace{-5mm}
  \label{ab_pv_cons}
\end{table}

\noindent\textbf{Design for UI2DPrompt.} UI2DPrompt consists of prompt construction and a hybrid injection module. Table~\ref{ab_pv_cons} shows the strategy selection for constructing PV prompts. Compared to a depth-based 3D transition, IPM is more suitable for coordinating the transformation based on the assumption of a flattened road surface. When uncertainty is applied both as dynamic weighting and as learning-based embeddings, performance is further enhanced by reliable PV prompts, achieving an improvement of +2.2 mAP.

Table~\ref{ab_hybridatt} demonstrates the impact of the hybrid injection design, with a total performance of 39.8 mAP for combining prompt-to-BEV and prompt-to-query attention. This result confirms the optimal utilization of reliable structural information, effectively compensating for information loss and errors introduced in the basic learning process.

\begin{table}[ht!]
  \centering
  \renewcommand\tabcolsep{5.pt}
   \footnotesize
  \begin{tabular}{ cc|cc c c}
    \toprule[1pt]
    
    \textbf{P2BEV}& \textbf{P2Q} & \textbf{AP}$_{\mathbf{ped}}$ & \textbf{AP}$_{\mathbf{div}}$ & \textbf{AP}$_{\mathbf{bou}}$ & \textbf{mAP} \\ 
    \midrule
     &   & 39.3 & 30.4 & 45.5 &  38.4   \\
     \cmark&     & 39.9 & 31.1 & 46.0 & 39.2\\ 
     & \cmark   & 39.6 & 30.7  &46.1  &38.8 \\
     \rowcolor{blue!10}
   \cmark & \cmark & 40.3 & 31.8 & 47.2 & \textbf{39.8}   \\
    \bottomrule[1pt]
    \end{tabular}
\vspace{-2mm}
  \caption{Ablations on hybrid injection. P2BEV and P2Q mean the cross-attention for $\text{F}_{\text{bev}}$ and $Q_{\text{map}}$ separately.}
  \label{ab_hybridatt}
  \vspace{-3mm}
\end{table}

\begin{table}[ht!]
\centering
\renewcommand\tabcolsep{14pt}
  \renewcommand\tabcolsep{7.pt}
   \footnotesize
  \begin{tabular}{c|cc>{\columncolor{blue!10}}c c| c}
    \toprule[1pt]
    $\boldsymbol{\lambda}_{\textbf{distill}}$ &5& 7 & 10 & 15 &\textcolor{gray}{/}\\ 
    \midrule
    \textbf{mAP} & 37.5 & 38.2 &  \textbf{39.3} & 38.8  &\textcolor{gray}{39.8}  \\
    \bottomrule[1pt]
    \end{tabular}
 \vspace{-2mm}
    \caption{Ablations on $\lambda_{\text{distill}}$ for MQ-Distillation Design. The last column refers to the use of the best teacher of PV prompts.}
    \vspace{-5mm}
  \label{ab_distill_loss}
\end{table}
\noindent\textbf{Ablations on MQ-Distillation.} To verify the effectiveness of the distillation module, we conduct ablations on the hyperparameter of distillation loss weight $\lambda_{\text{distill}}$, as this affects the imitation learning of PV structural features. As illustrated in Table~\ref{ab_distill_loss}, the best performance is achieved by setting the weight to 10.0, which refers to the best learning process of the mimic query for explicit structural features. 
\section{Conclusion and Future Work}

In this paper, we propose UIGenMap, an uncertainty-instructed structure injection approach to achieve a more generalized HD map construction. Specifically, we introduce the PV detection branch to obtain explicit structural features, in which the UA-Decoder produces adaptive probabilistic outputs. The UI2DPrompt module is proposed to construct reliable PV prompts and achieve dynamic integration through the hybrid injection designed. For real-time inference, MQ-Distillation is designed to mimic the structural prompts of PV.
Comprehensive experiments on the challenging geo-based partitions of nuScenes and Argoverse2 datasets demonstrate the generalization of our approach.

\noindent\textbf{For limitations}, current methods are designed and evaluated primarily on specific datasets. In the future, developing larger and more diverse datasets and benchmarks across various locations remains a crucial area for exploration. 

\section*{Acknowledgments}
This work is supported by National Natural Science Foundation of China under Grant No.62376244. It is also supported by Information Technology Center and State Key Lab of CAD\&CG, Zhejiang University.

{
    \small
    \bibliographystyle{ieeenat_fullname}

}

\appendix
\maketitlesupplementary

\setcounter{figure}{0}
\setcounter{table}{0}
\renewcommand{\thefigure}{A\arabic{figure}}
\renewcommand{\thetable}{A\arabic{table}}

\noindent In this supplementary material, we provide more details on our implementation and experiments as follows. 
\begin{itemize}
 \item Section \ref{sec:dataset}: More details on dataset partitions;
    \item Section \ref{sec:imple}: More details on implementation;
    \item Section \ref{sec:exps}: Additional experiments;
    \item Section \ref{sec:vis}: Qualitative results and failure case analysis;
    \item Section \ref{sec:limit}: Limitations and future work.
\end{itemize}

\section{Dataset Partitions}
\label{sec:dataset}
To evaluate the generalization capability for online HD map vectorization, we conducted experiments on the geospatial disjoint (geo-based) dataset partitions. Here are some statistical analyses of the datasets. Table~\ref{tab:stat} shows the number of samples and overlap ratios for the original and region-based data partitions. Compared to widely-used original splits, region-based partition divides datasets according to the geographic location with lower overlap rates, with $11\%$ for nuScenes~\cite{nuScenes} and $0 \%$ overlap for Argoverse2~\cite{Argoverse2}, which are measured for all locations within a 30m radius around the ego-vehicles. For the city-based partition following~\cite{lilja2024localization}, Table~\ref{city} shows the city-based distribution of the training and validation data. All dataset partitions are maintained on roughly the same scale as the original ones, in which Argoverse2 is resampled by 1/5 (2Hz) to be consistent with nuScenes dataset. 

\begin{table}[ht!]
  \centering
  \footnotesize
   \renewcommand\tabcolsep{6pt}
    \renewcommand\arraystretch{1.4}
    \begin{tabular}{@{}c|c|ccc@{}}
    \toprule[1pt]
     & \textbf{Split} & \textbf{Train \#} & \textbf{Val. \#} & \textbf{Overlap Ratio} \\
    \midrule
    
    \multirow{2}{*}{nuScenes} & Ori. & 27968 & 6019 & 85\%
     \\
    \multirow{2}{*}{ } & Region & 27846 & 5981 &   \textbf{11\%}\\
    \midrule
    \multirow{2}{*}{Argoverse2} & Ori. & 21794  & 4704  &   54\%
     \\
    \multirow{2}{*}{ } & Region & 23434 & 4676 &  \textbf{0\%} \\
    \bottomrule[1pt]
  \end{tabular}
  \caption{Data collections on region-based partitions.}
  \label{tab:stat}
\end{table}

\begin{table}[ht!]
    \centering
    \footnotesize
     \renewcommand\tabcolsep{3pt}
    \renewcommand\arraystretch{1.6}
    \begin{tabular}{c|cc|cc}
    \toprule[1pt]
     & \multicolumn{2}{c|}{\textbf{nuScenes}} &\multicolumn{2}{c}{\textbf{Argoverse2}}\\
                 & \textbf{City} & \textbf{Sample \#} & \textbf{City} & \textbf{Sample \#} \\
        \midrule
        Train &\makecell{Boston, \\Onenorth} & 25926 &\makecell{Miami, \\Pittsburgh}& 21975\\
        \midrule
        Val& \makecell{Queentown, \\Holland Village}& 8056 &\makecell{Austin, Detroit
    \\Washington} & 9232\\
    \bottomrule[1pt]
    \end{tabular}
    \caption{Data collections on city-based partitions.}
    \label{city}
\end{table}

\section{More Implementation Details}
\label{sec:imple}
This section introduces the detailed settings of pre-trained perspective-view (PV) detection branches, including the training process and the transformation of the ground truth of PV-level detection.
\subsection{Settings of PV Detection Branch} As described in the main paper, we introduce the pre-trained PV detection branch for explicit structural priors. In the design of the PV branch, given the total number of $I$ images captured by the onboard cameras, we utilize the shared ResNet50~\cite{he2016deep} backbone followed by the FPN~\cite{lin2017feature} neck to extract multi-scale image features. Then, a multi-layer uncertainty-aware (UA) decoder is employed to detect map elements, in which we replace the original Deformable-DETR~\cite{zhu2020deformable} with the UA-decoder design to take consideration of reliable PV coordinate output and uncertainty information, which will be further deployed in our UI2DPrompt module. We trained the model on four A100 GPUs with a batch size of $4$, which refers to $4\times I$ images in one batch. The learning rate is set to $3e^{-4}$ and the PV models are trained for $24$ epochs. The loss setting is similar to the main branch of the BEV map, which contains $\mathcal{L}^{\text{pv}}_{\text{nll}}$ fused with the loss of the point regression $\mathcal{L}^{\text{pv}}_{\text{pts}}$ to produce uncertainty of the Laplace distribution and stabilize the coordinate output, in addition to $\mathcal{L}^{\text{pv}}_{\text{cls}}$ for classification.
Thus, the PV loss can be formulated as below:
\begin{equation}
    \mathcal{L}^{\text{pv}}_{\text{map}} = 
\lambda^{\text{pv}}_{\text{pts}}\mathcal{L}^{\text{pv}}_{\text{pts}} + 
\lambda^{\text{pv}}_{\text{nll}}\mathcal{L}^{\text{pv}}_{\text{nll}} + \lambda^{\text{pv}}_{\text{cls}}\mathcal{L}^{\text{pv}}_{\text{cls}},
\end{equation}
where the corresponding loss weights $\lambda^{\text{pv}}_{\text{nll}}$, $\lambda^{\text{pv}}_{\text{pts}}$, and $\lambda^{\text{pv}}_{\text{cls}}$ are set to $0.05$, $50.0$, and $5.0$, respectively.

\subsection{Ground Truth for PV Detection} Since the map annotations are labeled in the bird's-eye-view (BEV) space, the PV map labels are obtained through the projection of the BEV map ground truth. Given map ground truth at the ego-coordinate system $(p_x,p_y)$, map polylines are transformed into the image-coordinate system with 2D coordinates
$(x_\text{pv},y_\text{pv})$ by camera extrinsic $\text{T}_{\text{ego2cam}}$ and intrinsic $\text{K}_\text{cam}$, which can be formulated as: 
\begin{equation}
\mathbf{P}_{\text{cam}}^{\text{h}} = \text{T}_{\text{ego2cam}} \cdot \begin{bmatrix} p_x \\ p_y \\ 0 \\ 1 \end{bmatrix},
\end{equation}
\begin{equation}
\begin{bmatrix} x_\text{pv} \\ y_\text{pv} \\ 1 \end{bmatrix} =  \frac{1}{z_{\text{cam}}} \cdot\text{K}_\text{cam} \cdot \mathbf{P}_{\text{cam}}^{\text{3D}},
\end{equation}
where $\mathbf{P}_{\text{cam}}^{\text{3D}}$ is the first three dimension of $\mathbf{P}_{\text{cam}}^{\text{h}}$. The depth of the camera coordinates $z_\text{cam}$ is used for normalization when projecting a point from the 3D space (camera coordinate system) onto the 2D image plane. Furthermore, the corresponding PV map elements are cropped and filtered according to the depths and range of images.

\section{Additional Experiments}
\label{sec:exps}

\subsection{More Ablation Studies}
In this section, we demonstrate more ablations on the selection of hyper-parameters. All experiments are conducted with the utilization of mimic query distillation.

\noindent\textbf{{Ablations on the Uncertainty Loss Weight.}} In Table~\ref{ab_nll_loss}, we conduct ablations on the selection of $\lambda_{\text{nll}}$ for uncertainty head loss. It can be observed that the best performance is achieved when the hyperparameter $\lambda_{\text{nll}}$ is set to 0.05. Excessive or insufficient weights may cause imbalances in learning, thereby affecting the model's performance.
\begin{table}[th!]
\centering
  \renewcommand\arraystretch{1.1}
  \renewcommand\tabcolsep{6.pt}
   \small
  \begin{tabular}{ r|cc c c}
    \toprule[1pt]
    
    $\boldsymbol{\lambda}_{\textbf{nll}}$ & \textbf{AP}$_{\mathbf{ped}}$ & \textbf{AP}$_{\mathbf{div}}$ & \textbf{AP}$_{\mathbf{bou}}$ & \textbf{mAP} \\ 
    \midrule
     0.00 & 39.2&30.6&44.4&38.1\\
     0.02 &   40.7  & 30.7  & 45.0  &   38.8  \\
     \rowcolor{blue!10}
    0.05  &  40.3 & 30.8   &  46.8 & \textbf{39.3} \\
    0.07 & 40.1  & 29.8  & 45.1  & 38.9   \\
    0.10 &  39.5 &  30.1 & 45.3  &  38.3  \\
    \bottomrule[1pt]
    \end{tabular}
  \vspace{-2mm}
  \caption{Ablations on the $\lambda_{\text{nll}}$ of $L_{\text{nll}}$ for UA-Head output.}
    \vspace{-2mm}
  \label{ab_nll_loss}
\end{table}

\noindent\textbf{{Ablations on Threshold Selection.}}
To examine the selection scheme for PV instances, we conducted ablations on the various settings of $c_{\text{thr}}$ in our UI2DPrompt design. As an uncertainty output, a larger $c_{\text{thr}}$ indicates higher selection standards. Excessive or insufficient threshold selection may lead to the loss of critical features or the introduction of redundant information. As shown in Table~\ref{ab_thre}, the best performance of 39.3 mAP is obtained with $c_{\text{thr}}=0.4$. 

\begin{table}[th!]
\centering
\renewcommand\tabcolsep{9pt}
\footnotesize
  \begin{tabular}{c|c c cc}
    \toprule[1pt]
    $\boldsymbol{c}_{\textbf{thr}}$& 0.2 & \cellcolor{blue!10}0.4 &  0.5 &0.6\\ 
    \midrule
    \textbf{mAP} &  39.0 & \cellcolor{blue!10}\textbf{39.3}  &  38.8 &  38.6   \\
    \bottomrule[1pt]
    \end{tabular}
  \vspace{-2mm}
  \caption{Ablations on threshold selection.}
    \vspace{-4mm}
  \label{ab_thre}
\end{table}

\section{Visualization}
\label{sec:vis}
\subsection{Visual Comparisons}
\begin{figure*}
\begin{center}
\centering
\includegraphics[width=0.98\textwidth]{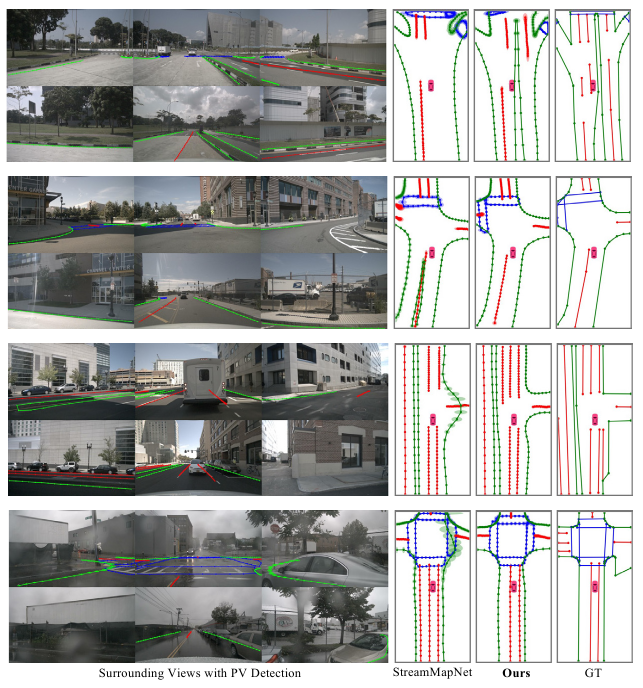}
\caption{Visual Comparisons on region-based nuScenes validation set with PV detection result. Circles represent the point-based uncertainty, larger circle means less confidence for model's prediction. Our approach achieves better performance with less uncertainty.}
\vspace{-4mm}
\label{fig_visual_nus1}
\end{center}
\end{figure*}

\begin{figure*}
\begin{center}
\centering
\includegraphics[width=0.98\textwidth]{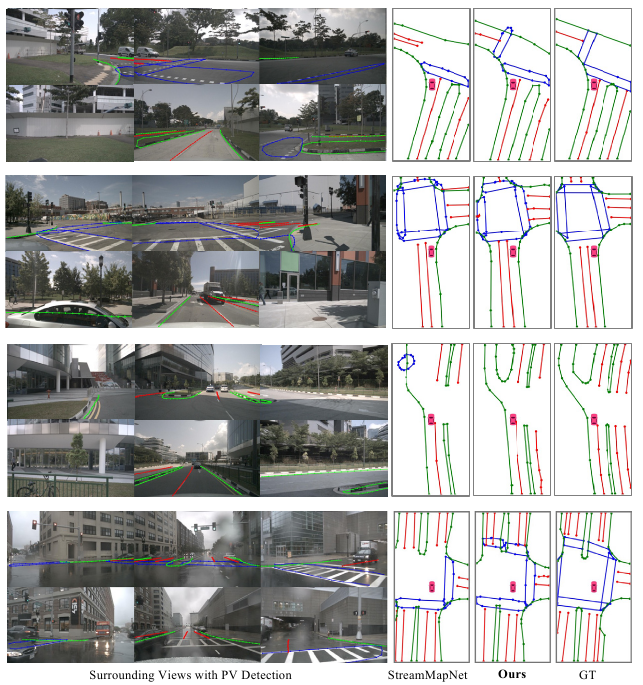}
\caption{Visual Comparisons on region-based nuScenes validation set with PV detection result, in which our approach achieves better performance. }
\vspace{-4mm}
\label{fig_visual_nus2}
\end{center}
\end{figure*}

\begin{figure*}
\begin{center}
\centering
\includegraphics[width=0.98\textwidth]{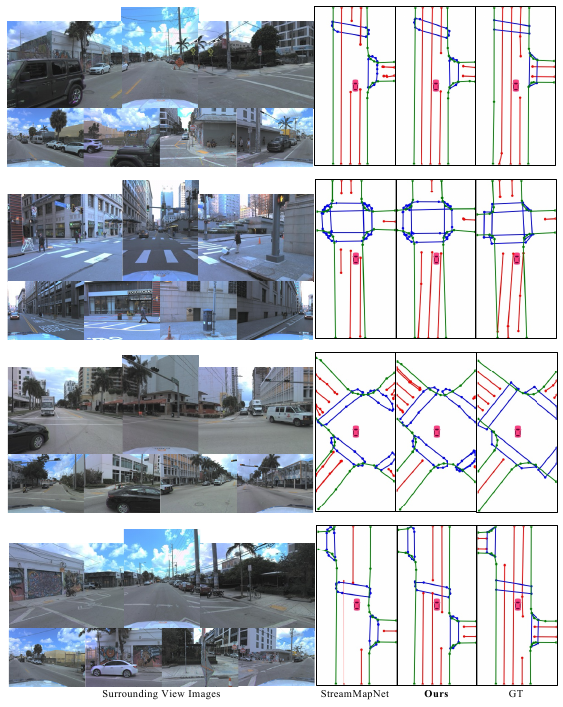}
\caption{Visual comparisons on region-based Argoverse2 validation dataset. Our UIGenMap emphasizes stronger generalized ability.}
\vspace{-4mm}
\label{fig_visual_argo1}
\end{center}
\end{figure*}

\begin{figure*}
\begin{center}
\centering
\includegraphics[width=0.98\textwidth]{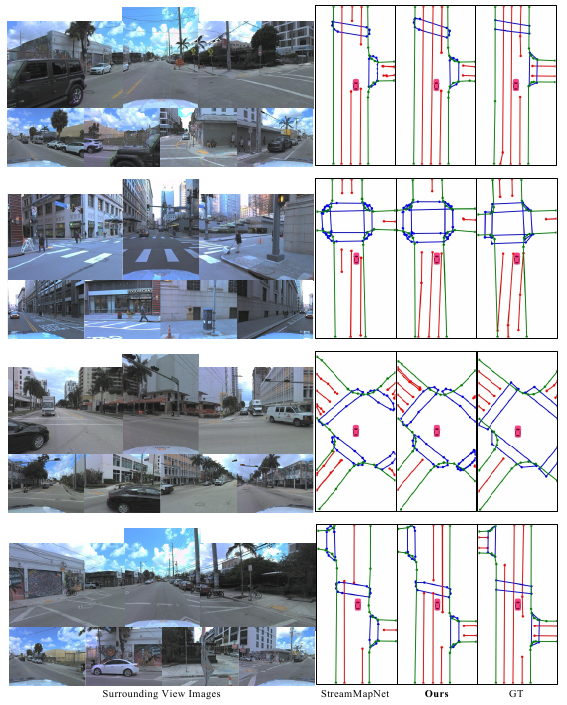}
\caption{More visual comparisons on region-based Argoverse2 validation dataset.}
\vspace{-4mm}
\label{fig_visual_argo2}
\end{center}
\end{figure*}

On region-based splits, Figure~\ref{fig_visual_nus1} and Figure~\ref{fig_visual_nus2} present additional visual comparisons on nuScenes validation set. In Figure~\ref{fig_visual_nus1}, uncertainties are presented as circles of different sizes. The larger circle represents a lower confidence in its prediction. As shown in the fifth sample, the larger uncertainty is observed in the \texttt{FRONT-RIGHT} view of the image, mainly due to occlusions caused by the car. Compared to the previous method, our UIGenMap performs better with lower uncertainties, particularly for road boundaries and pedestrian crossings.
Figure~\ref{fig_visual_argo1} and Figure~\ref{fig_visual_argo2} present more qualitative visual comparisons on the region-based Argoverse2 dataset.


\subsection{Failure Case Analysis}
\begin{figure*}
\begin{center}
\centering
\includegraphics[width=0.98\textwidth]{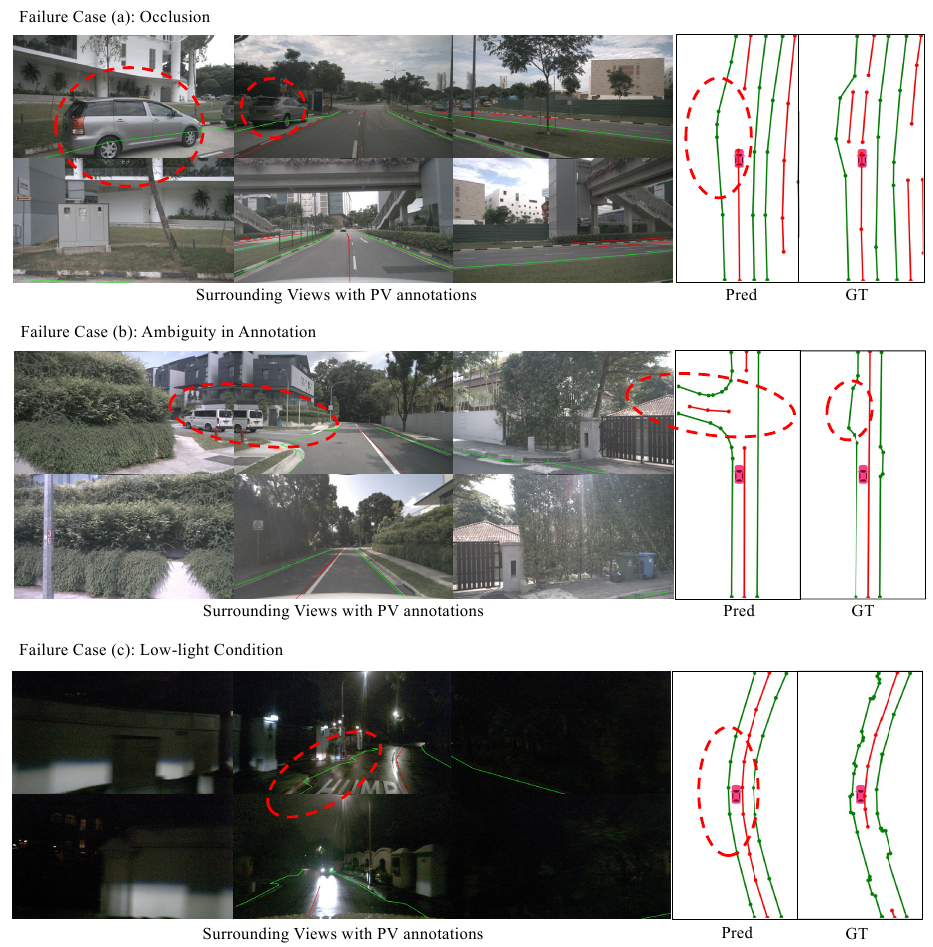}
\caption{Visualization of Failure cases. (a): Occlusion by dynamic objects; (b): Wrong calibration and annotations in publically used dataset; (c): Driving at night with low-light conditions.}
\vspace{-4mm}
\label{fig_failue2}
\end{center}
\end{figure*}

Despite having greatly improved the quality of generalizable HD map construction, both the visual and numerical results show that there is still a large gap in the requirement for real-world deployment. In Figure~\ref{fig_failue2}, we provide some visual examples of failure cases. 

\noindent{\textbf{Occlusion.}}
As shown in Figure~\ref{fig_failue2} (a), static map elements can be repeatedly occluded by dynamic objects on the road, which may cause a limited field of view and inadequate detection of key elements of the road.

\noindent{\textbf{Ambiguous Annotations.}}
Figure~\ref{fig_failue2} (b) presents an example of annotation errors. A left-turn junction can be seen on the front- and front-left-view of the PV images, which is unlabeled in the ground truth.

\noindent{\textbf{Low-light Conditions.}}
For night driving and other low-light conditions, PV images cannot provide enough semantic and structural information. As shown in Figure~\ref{fig_failue2} (c), it is hard to capture detailed road structures, so there is potential for further enhancement.

\section{Limitations and Future Work}
\label{sec:limit}
\noindent{\textbf{Limitations.}} Considering the generalization capability, the performance of learned models is heavily affected by the scale and diversity of the training data. In this paper, experiments are conducted within the same dataset. So, there is a lack of extension to generalization studies across different datasets. More strategies like modality fusion, data augmentation, and different modeling strategies for map element representation can be utilized for stronger generalization capability. Within the limited training data, performance on unseen driving scenarios remains constrained, which poses significant challenges for deployment in real-world applications. 

\noindent{\textbf{Future work.}} In the future, further explorations are required to improve the generalization of map construction, particularly under adverse conditions such as rain, clouds, and fog. In addition, the impact of different sensor models and placements on generalization must be addressed. For practical industrial applications, it is crucial to develop larger datasets and benchmarks that include a wide range of locations and scenarios. These resources would enable models to better manage the variability and complexity of real-world driving conditions. 

\clearpage

\end{document}